\documentclass[10pt,twocolumn,letterpaper]{article}

\usepackage{iccv}
\usepackage{times}
\usepackage{epsfig}
\usepackage{graphicx}
\usepackage{amsmath}
\usepackage{amssymb}
\usepackage{epsfig}
\usepackage{graphicx}
\usepackage{amsmath}
\usepackage{tabulary}
\usepackage{stfloats}
\usepackage{amssymb}
\usepackage[table,xcdraw]{xcolor}
\usepackage{multirow}
\usepackage{color, colortbl}
\usepackage{wrapfig}
\usepackage{pdfpages}
\usepackage{enumerate}
\usepackage{footnote}
\usepackage{multirow}
\usepackage{makecell}
\usepackage{subfig}

\usepackage[pagebackref=true,breaklinks=true,letterpaper=true,colorlinks,bookmarks=false]{hyperref}

\iccvfinalcopy 

\def\iccvPaperID{2179} 

\ificcvfinal\fi

\begin{document}

\title{Constructing Self-motivated Pyramid Curriculums for Cross-Domain Semantic Segmentation: A Non-Adversarial Approach}
\author{Qing Lian\textsuperscript{1},~~~  Fengmao Lv\textsuperscript{1},~~~Lixin Duan\textsuperscript{1},~~~ Boqing Gong\textsuperscript{2} \\
\textsuperscript{1}University of Electronic Science and Technology of China~~~ \textsuperscript{2}Google \\
{\tt\small \{lianqing1997, lxduan\}@gmail.com, fengmaolv@126.com, boqinggo@outlook.com}
}

\maketitle
\ificcvfinal\thispagestyle{empty}\fi

\begin{abstract}

We propose a new approach, called self-motivated pyramid curriculum domain adaptation (PyCDA), to facilitate the adaptation of semantic segmentation neural networks from synthetic source domains to real target domains. Our approach draws on an  insight connecting two existing works: curriculum domain adaptation and self-training. Inspired by the former, PyCDA constructs a pyramid curriculum which contains various properties about the target domain. Those properties are mainly about the desired label distributions over the target domain images, image regions, and pixels. By enforcing the segmentation neural network to observe those properties, we can improve the network's generalization capability to the target domain. Motivated by the self-training, we infer this pyramid of  properties  by resorting to the semantic segmentation network itself. Unlike  prior work, we do not need to maintain any additional models (e.g., logistic regression or discriminator networks) or to solve  $\min\max$ problems which are often difficult to optimize. We report state-of-the-art results for the adaptation from both GTAV and SYNTHIA to Cityscapes, two popular settings in unsupervised domain adaptation for semantic segmentation.
\end{abstract}

\section{Introduction}

\begin{figure}%
\centering
\subfloat[Synthetic images with labeling-free pixel-wise groundtruth annotations.]{%
\label{fig:fig1a}%
\includegraphics[width=1.0\columnwidth]{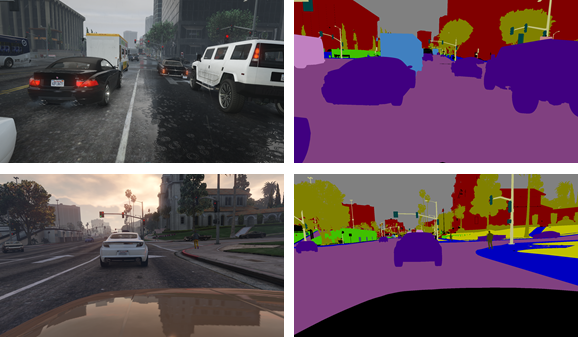}
}%
\qquad
\subfloat[Segmentation results of a real image with and without adaptation.]{%
\label{fig:fig1b}%
\includegraphics[width=1.0\columnwidth]{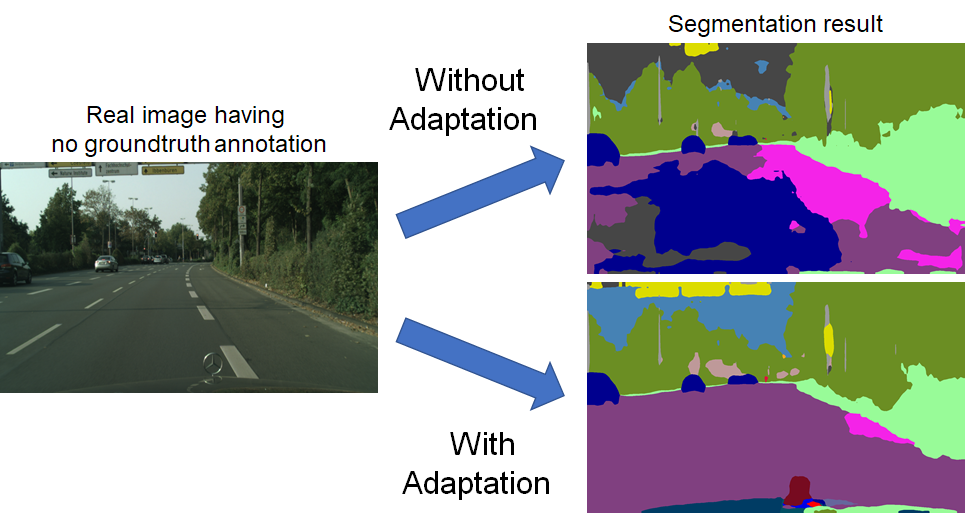}
}%
\vspace{-0.1cm}
\caption{Unsupervised domain adaptation for  semantic segmentation. The segmentation results for real images can be significantly improved by explicit domain adaptation techniques when we  adapt a segmentation model trained using synthetic imagery.}\label{fig:fig1}
\vspace{-0.5cm}
\end{figure}

The objective of semantic segmentation is to assign a semantic label to every pixel of an image. Over the past few years, a great amount of effort has been made by the community to tackle this problem~\cite{Long_2015_fcn, deeplab, zhao2017pspnet, deeplabv3plus2018, RefineNet}, leading to sophisticated and high-performing deep convolutional neural networks as the main  solution. However, to collect and label images for training such networks is a very daunting work~\cite{Cordts2016Cityscapes}.  To alleviate the heavy annotation burden, a promising alternative is to employ photo-realistic simulators to efficiently collect and label training data. Richter et al.~\cite{gtav_2016_ECCV} use the GTAV game engine to aid user annotations, resulting a dataset of 25 thousand synthetic urban scene images labeled in only 49 hours (about 7 seconds per image).

The clear visual mismatch  between the synthetic (source) domains and  real (target) domains (cf.~Figs.~\ref{fig:synthetic-datasets} and~\ref{fig:qualitative}), however, inevitably causes significant performance degradation when one applies the model trained on the source domain to the real images of the target domain.

In order to better take advantage of the synthetic imagery for the semantic segmentation of real scenes, we propose a novel domain adaptation approach by drawing an intriguing connection between two previous works: curriculum domain adaptation~\cite{zhang2017curriculum,curriculum_tpami} and self-training~\cite{self_motivated_2018_ECCV}. This connection naturally leads to self-motivated pyramid curriculums and a new training algorithm for the cross-domain adaptation of  semantic segmentation networks. Compared with the prevalent adversarial training methods in domain adaptation~\cite{road_2018_CVPR,fcn_wild,fcan_2018_CVPR, Adaptseg_Tsai_2018_CVPR,I2I_Murez_2018_CVPR, No_Discrimination_2017_ICCV, Hoffman_cycada2017, CGA_Hong_2018_CVPR, Sankaranarayanan_2018_CVPR}, our approach gives rise to results on par with or better than theirs and yet is lighter-weight, without the need of learning extra discriminator nets, and easier to optimize, without the need of solving any $\min\max$ problems. More importantly, it outperforms both the original curriculum adaptation~\cite{curriculum_tpami} and the original self-training method~\cite{self_motivated_2018_ECCV}.


In particular, we view self-training from the perspective of curriculum domain adaptation. They share the same algorithmic format. On the one hand, the self-training alternates between two sub-tasks: 1) estimating pseudo labels for the target domain's pixels and 2) updating the weights of the segmentation network by using both the source labels and the pseudo target labels. On the other hand, the curriculum adaptation first 1) constructs a curriculum, i.e., infers properties of the target domain in the form of frequency distributions of the class labels over an image (or image region) and then 2) updates the network's weights using the source labels and the target domain's properties. Due to this analogy, we may view the pseudo labels in self-training as one of the properties about the target domain. More interestingly, the second steps of the two works share exactly the same form in math --- a cross-entropy loss between a frequency distribution / pseudo label and a differentiable function of the network's predictions.

Immediately, our approach follows the above analogy. We add the pseudo labels in self-training to the curriculum as the finest layer of properties about the target domain images. On top of that, we build a pyramid of which a layer comprises image regions of a certain size. This pyramid design resembles the original curriculum domain adaptation work, in which the frequency distributions are counted over a global image and some superpixels, respectively --- in other words, a simple two-layer pyramid.

In addition to enriching the original curriculum by the pseudo labels, we also improve it in two ways. One is to replace the superpixels by small squared regions to significantly save computation cost. The other is to infer the target domain properties --- label distributions over the squared regions and full target domain images --- by the semantic segmentation network itself. In each  iteration of the training phase, we infer those  properties from the network's pxiel-wise predictions over the target domain images and then use a loss defined over those properties to update the network by backpropagation.


Our main contribution is two-fold. One is that we provide a new insight connecting the self-training for adapting segmentation networks~\cite{self_motivated_2018_ECCV} and the curriculum adaptation method~\cite{zhang2017curriculum,curriculum_tpami}.  The other is that, inspired by the connection, we propose a novel self-motivated pyramid curriculum for the domain adaptation of semantic segmentation networks. Extensive experiments show that it outperforms either~\cite{self_motivated_2018_ECCV}~or~\cite{curriculum_tpami} individually. Moreover, it is on par with or better than state-of-the-art adversarial adaptation methods without the need of maintaining an extra discrminator network or carefully tuning the optimization procedure for $\min\max$ problems.

\section{Related Work}
\noindent{{\bf Semantic segmentation.}}
Semantic segmentation is the task that assigns labels in pixel level for an image which plays a vital role in lots of tasks including autonomous driving, disease detection, etc. In the following, we briefly review some of the works with a focus on CNN-based methods. Driven by the powerful deep neural networks~\cite{He_2016_resnet}, pixel-level prediction tasks achieve great progress mostly following the design of replacing the softmax layer in classification with the pixel-wise softmax~\cite{Long_2015_fcn}. To enlarge the receptive fields and feature resolutions, methods of~\cite{deeplab, chen2017rethinking,deeplabv3plus2018, zhao2017pspnet} employ dilated convolution. To utilize different context information and multiple features, some extend the dilated convolution to a pyramid~\cite{deeplab,chen2017rethinking,deeplabv3plus2018} or resize it with multiple sizes~\cite{zhao2017pspnet}.

\vspace{3pt}
\noindent{{\bf Domain adaptation.}}
A basic assumption in conventional machine learning is that the training and test data are drawn i.i.d.\ from the same underlying distribution. However, this does not always hold in real world scenarios, resulting in significant performance drops when the training and test data exist distribution mismatches. Domain adaptation aims to rectify this mismatch and make the model generalize well onto the test domain. Domain adaptation has been mostly addressed for image classification problems in computer vision~\cite{DuanTPAMI2012a,DuanTPAMI2012b,gong_gfk,gong_landmarks,dann,dan,jan}. Recent works start to study deep neural networks including learning domain-invariant models~\cite{dann,MCD_Saito_2018_CVPR,gan2016learning} and target specific models~\cite{dirt_t, french2018selfensembling}.

\vspace{3pt}
\noindent{{\bf Domain adaptation for semantic segmentation.}}
Recent work on domain adaptation mainly proceeds in two directions. One is based on a curriculum learning strategy. Zhang et al.\ first learns to solve easy tasks in the target domain and then use them to regularize semantic segmentation~\cite{zhang2017curriculum,curriculum_tpami}. Dai et al.\ construct a curriculum by simulating foggy images of different fog densities~\cite{dai2019curriculum}. In this work, we also view the self-training~\cite{self_motivated_2018_ECCV} as a curriculum-style domain adaptation method. The other line of work is to reduce the domain shift in feature space or output space and tries to seek a better way to align both domains in an intermediate layer~\cite{road_2018_CVPR,fcn_wild,fcan_2018_CVPR, Adaptseg_Tsai_2018_CVPR,I2I_Murez_2018_CVPR, Penalize_top_2018_ECCV, No_Discrimination_2017_ICCV, Hoffman_cycada2017, CGA_Hong_2018_CVPR, Sankaranarayanan_2018_CVPR,luo2018taking}. In~\cite{fcn_wild}, Hoffman et al.\ combine global and local alignment methods with a domain adversarial training loss. In~\cite{No_Discrimination_2017_ICCV}, Chen et al.\ further propose a global and class-specific feature alignment approach guided by the soft pseudo labels in the target domain. In~\cite{Adaptseg_Tsai_2018_CVPR}, Tsai et al.\ proposed to align both domains at the structured output space level. In~\cite{road_2018_CVPR, fcan_2018_CVPR}, they focus on utilizing spatial information to help the discriminator distinguish the domains better. In~\cite{luo2018taking, Penalize_top_2018_ECCV}, they focus on solving the problem of class boundary or outliers in the domain adaptation. In~\cite{Hoffman_cycada2017, I2I_Murez_2018_CVPR, Sankaranarayanan_2018_CVPR}, they propose to learn domain adaptive segmentation networks through directly translating the source images to the target ones at the pixel level. We refer readers to \cite[Section 5]{curriculum_tpami} for a more thorough review.

\section{Approach}\label{sec:Approach}

\begin{figure*}[]
	\centering
    \includegraphics[width=0.83\textwidth]{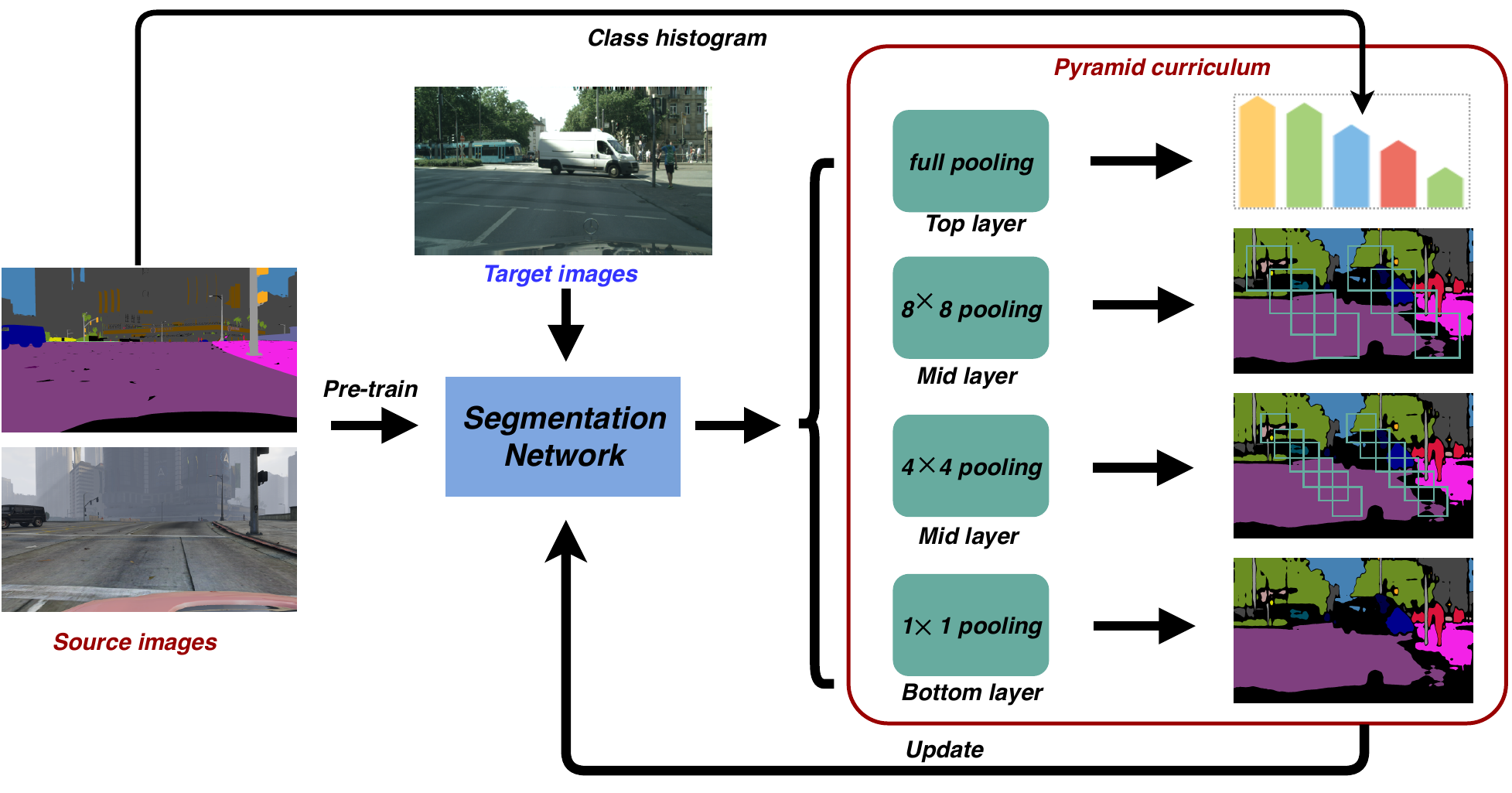}
    \vspace{-1mm}
	\caption{Overview of our self-motivated pyramid curriculum domain adaptation (PyCDA) approach to segmentation.}
	\label{fig:framework}
	\vspace{-4mm}
\end{figure*}

In this section, we first reveal the connection between the curriculum domain adaptation (CDA)~\cite{curriculum_tpami} and the self-training (ST)  for adaptation~\cite{self_motivated_2018_ECCV}. This connection naturally leads to the training algorithm of this paper, dubbed self-motivated pyramid curriculum domain adaptation (PyCDA), for the semantic segmentation task.

\subsection{CDA \textit{vs.}\ ST}
Denote by $I_t\in \mathbb{R}^{H\times W}$  a target domain image and $\hat{Y}_t\in\mathbb{R}^{H\times W\times C}$  the corresponding output of a semantic segmentation network, where $H$, $W$, and $C$ are respectively the height,  width, and number of possible classes of the input image. Most neural segmentation networks employ a pixel-wise softmax function at the output layer so that $\hat{Y}_t(i,j)\in\mathbb{R}^C$ is a probability vector satisfying $\sum_c \hat{Y}_t(i,j,c)=1, \forall i,j$ and $\hat{Y}_t(i,j,c)\ge 0, \forall i, j, c$. Similarly, denote by $I_s$ and $\hat{Y}_s$ a source domain image and the corresponding prediction by a network. In unsupervised domain adaptation, the groundtruth labels $\{Y_s\}$ of the source domain ($S$) are given but the learner has no access to the labels of the target domain ($T$).
\vspace{3pt}

\noindent{{\bf CDA}}~\cite{curriculum_tpami}
 learns a semantic segmentation network by minimizing the following objective function:
\begin{align}
    \min \quad \sum_{s\in S}\mathcal{L}(Y_s, \hat{Y}_s) + \lambda \sum_{t\in T}\sum_{k\in \mathcal{P}_t^1} \mathcal{C}(p_t^k, \hat{p}_t^k)
    \label{eq1}
\end{align}
where the first term sums up pixel-wise cross-entropy losses over the source domain images ($s\in S$), each summand of the second term is a cross-entropy loss over a target domain image ($t\in T$) between two label distributions which indicate the proportion of each class in the image $t$ or in a region of it, and the set $\mathcal{P}_t^1$ collects all the label distributions for the target image $t$. As a  concrete example, a desired label distribution ${p}_t^0$ over a target domain image $I_t$ is calculated by
\begin{align}
    p_t^0(c) = \frac{1}{WH}\sum_{i=1}^W\sum_{j=1}^H Y_t(i,j,c), \quad \forall c, \label{eP}
\end{align}
and the property $\hat{p}_t^0$ predicted by the segmentation network, which can be obtained in a similar way as above from the network prediction $\hat{Y}_t$, is supposed to match the desired label distribution $p_t^0$. In this example, the label distribution over a full image captures a \emph{global} property. The others calculated within an image region capture the \emph{local} properties of a target domain image. Accurate readers may wonder how to estimate the desired properties $p_t^k, k\in\mathcal{P}_t^1$, in practice because the target labels are actually unknown in unsupervised domain adaptation; we explain it below.

The name of CDA attributes to the following easy-to-difficult curriculum. Compared with the pixel-wise predictions, it is relatively easy to obtain these label distributions, especially when the images are about urban scenes which share common objects and spatial layouts. Zhang et al.~\cite{curriculum_tpami} train separate logistic regression models and support vector machines to estimate these distributions $p_t^k, k\in \mathcal{P}_t^1$. In this paper, however, we estimate them in a self-training fashion by the segmentation network itself. 

\vspace{3pt}
\noindent{{\bf Self-training (ST)}}~\cite{self_motivated_2018_ECCV} considers the unknown labels of the target domain images as latent variables. It alternates between 1) inferring the values of the latent target labels and 2) updating the network's weights. The second step essentially solves the following problem:
\begin{align}
    \min\; \sum_{s\in S}\mathcal{L}(Y_s,\hat{Y}_s) + \lambda\sum_{t\in T}\sum_{(i,j)\in \mathcal{P}_t^2} \mathcal{C}(Y_t{(i,j)}, \hat{Y}_t(i,j)), \notag
\end{align}
where $\mathcal{P}_t^2$ is the set of pixels of the target image $I_t$ whose pseudo labels $\{Y_t{(i,j)}\}$ are inferred --- the other pixels are with null labels because the network predictions at those positions are probably below a certain threshold.

\vspace{3pt}
\noindent{{\bf Connection between the two.}}
The objective functions of CDA and ST are remarkably alike. The only difference between them is on the sets $\mathcal{P}_t^1$ and $\mathcal{P}_t^2$, where the former collects label distributions over the global target domain image $I_t$ and some of its local regions, and the latter is the pixels of $I_t$ that are pseudo-labeled by the first step in ST. \emph{What if we union the two sets, i.e., $\mathcal{P}_t^1\cup\mathcal{P}_t^2$?} They indeed seem like mutually complementary. Thanks to CDA, we conjecture that the finest-grained pixel-level pseudo labels $P_t^2$ may be enhanced by the label distributions of the fine-grained image regions and the coarsest-grained full image. Due to ST, we reasonably expect the label distributions may be derived from the segmentation network itself, without the need of resorting to any additional models (e.g., logistic regression or SVM used in~\cite{curriculum_tpami}). Following this line of reasoning, we devise our approach as the following.

\subsection{Self-motivated pyramid CDA (PyCDA)} \label{sec-pyCDA}
We propose a self-motivated pyramid curriculum for the domain adaptation (PyCDA) of semantic segmentation tasks.  The idea faithfully follows the insight above, i.e., we union the two sets $\mathcal{P}_t^1\cup \mathcal{P}_t^2$ used in CDA and ST, respectively. This union results in a pyramid with at least three layers: pixels on the bottom layer, small image regions in the middle, and a full image on the top. {Note that this pyramid is built for a target domain image, not source domain images.} Similarly to CDA, we will infer the label distribution over the top-layer full image and label distributions, or more concretely one-hot vectors, for the middle-layer small image regions. Similarly to ST, we want to assign pseudo labels to some of the bottom-layer pixels. Please refer to Fig.~\ref{fig:framework} for an overview of the proposed PyCDA approach.

\vspace{3pt}
\noindent{{\bf Superpixels \emph{vs.}\ pixel squares.}}
Before describing the main approach, we first discuss how to partition an image into small regions. Zhang et al.\ employ in their original CDA work~\cite{curriculum_tpami} non-overlapping superpixels, which incur additional computation overhead. We replace the superpixels by overlapped squares instead. While the pixel squares do not track object boundaries, their squared shape enables fast GPU computation as demonstrated beblow when we infer the label distributions for them. Moreover, as the squares are sufficiently small, most of them each cover pixels of the same class.  In the experiments, we use the squares of $4\times4$ and $8\times8$ for the middle layers of the pyramid in PyCDA. 

\vspace{3pt}
\noindent{{\bf Self-motivated inference of target domain properties.}}
In order to estimate the pseudo label $Y_t{(i,j)}$ of a target domain pixel $(i,j)$, we use as simple as a thresholding method. Denoting by $c^\star\leftarrow\arg\max_c \hat{Y}_t(i,j,c)$, we have
\begin{align}
    Y_{t}{(i,j)} =
    \begin{cases}
    c^\star  & \text{if } \hat{Y}_t(i,j,c^\star)>0.5\\
    \text{null} & \text{otherwise}
    \end{cases} \label{eq:pseudo}
\end{align}
\vspace{-0.3cm}

\noindent where $\hat{Y}_t$ is the output of a segmentation network at a training iteration. We say a pixel survives this step and will add it to the bottom layer of the pyramid curriculum, if its pseudo label is not null. Alternatively, one may use the self-paced policy design~\cite{self_motivated_2018_ECCV} to estimate the pseudo labels.

We employ a similar strategy to decide to which class each pixel square belongs to. Denote by $(i_0,j_0)$ a  square (e.g., the coordinates of the top-left corner of this square). We take an average pooling of the network's predictions
\begin{align}
    \hat{Y}_\texttt{square}(i_0,j_0,c)\leftarrow\texttt{mean}_{(i,j)\in\texttt{square}}(\hat{Y}_t(i,j,c))，
\end{align}
and then threshold the pooled value $\hat{Y}_\texttt{square}(i_0,j_0,c)$ to determine the label of the square, i.e., by replacing $\hat{Y}_t$ with $\hat{Y}_\texttt{square}$ in eq.~(\ref{eq:pseudo}). In implementation, we leverage efficient GPU operations by adding an average pooling layer after the network's output layer. This label for a pixel square can be converted to a one-hot vector and regarded as a label distribution over this square.

Finally, we want to estimate the label distribution over a full target domain image. Zhang et al.~\cite{curriculum_tpami} gives a few candidate methods for this sub-task, and we use the most computation-efficient one in this work. Particularly, no matter for which target domain image, we transfer to it the mean of the label distributions of all source domain images.  As shown in the the experiments of~\cite{curriculum_tpami}, this actually gives rise to results on par with learning a logistic regression model from the source domain probably because the images of urban scenes share common objects and layouts. For the target domains beyond urban scenes, alternative sophisticated algorithms are desired to reliably estimate the label distributions for the target domain images.

\vspace{3pt}
\noindent{{\bf PyCDA.}}
We are now ready to present the overall objective function of the proposed PyCDA approach:
\vspace{-0.2cm}
\begin{align}
    \min\quad &\frac{1}{|S|}\sum_{s\in S}\mathcal{L}(Y_s,\hat{Y}_s) + \frac{\lambda_1}{|T|}\sum_{t\in T} \mathcal{C}\big(p_t^0,\hat{p}_t^0\big) \notag \\
    &+ \frac{\lambda_2}{|\mathcal{P}|}\sum_{(t,k)\in \mathcal{P}} \mathcal{C}\big(Y_t^k, \hat{Y}_t^k\big),
    \label{eq:pycda}
\end{align}
where $\lambda_1$ and $\lambda_2$ are pre-defined trade-off parameters, and $\mathcal{P}=\{(t,k)|\forall t\in T, k\in\mathcal{P}_t^1\cup\mathcal{P}_t^2\}$ denotes the label distributions (resp., pseudo labels) over the squared regions (resp., pixels) of the target images. Note that in Eq.~(\ref{eq:pycda}), the second term is a cross-entropy loss defined on the label distributions over target images, and the last term takes care of the pixels and squared regions of the target domain. In the experiments, we set $\lambda_1=1$, $\lambda_2=0.5$ and tune other free parameters (e.g., learning rate) via a validation set.

\vspace{3pt}
\noindent{{\bf Effect of the pyramid curriculum.}}
We have described a pyramid curriculum which consists of the global target domain images on the top, pixels at the bottom, and pixel squares in between. From the CDA point of view~\cite{curriculum_tpami}, the label distributions over the top-layer images macroscopically hint the network \emph{how} to update its predictions while the label distributions over the middle-layer
 pixel squares microscopically indicate the network \emph{where} to update. The pseudo labels of bottom-layer pixels give even more precise supervision to the network. From the ST perspective~\cite{self_motivated_2018_ECCV}, the middle-layer pixel squares may be viewed as a way of consensus voting so that the pseudo labels (of the squares) can be estimated more reliably than thresholding the prediction at an isolated pixel. The label distributions over the target domain images act like a prior over the classes, playing a similar role to the class-balanced formulation in~\cite{self_motivated_2018_ECCV}.

\section{Experiments}
\label{sec:dataset}

\begin{figure*}[ht]
\centering
\subfloat[GTAV]{%
\label{fig:fig3a}%
\includegraphics[width=1.0\linewidth]{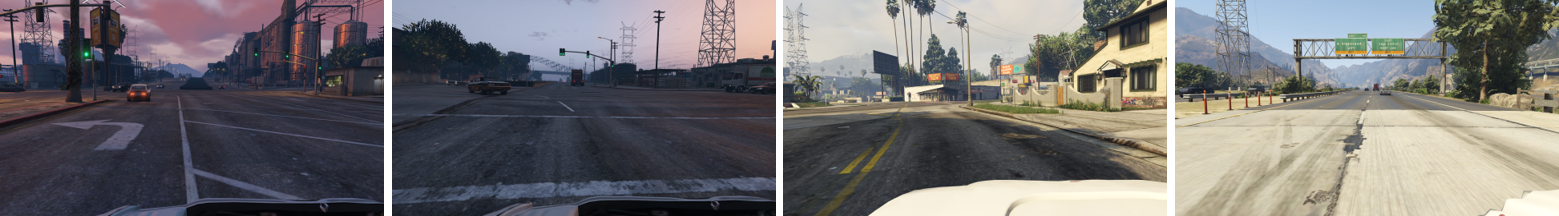}
}%
\vspace{-0.4cm}
\qquad
\vspace{-0.2cm}
\subfloat[SYNTHIA]{%
\label{fig:fig3b}%
\includegraphics[width=1.0\linewidth]{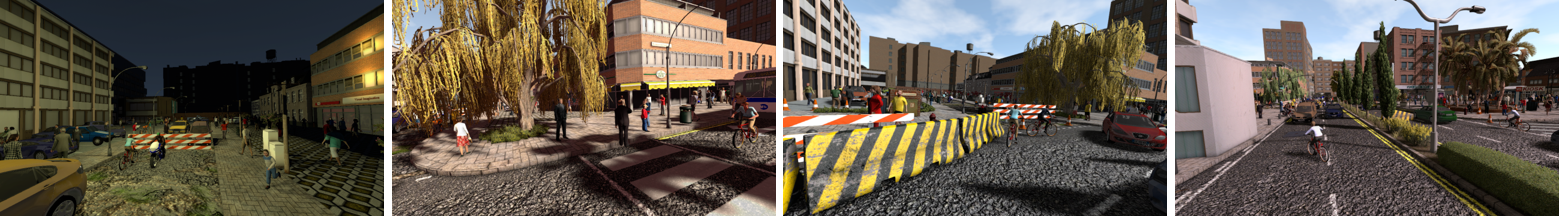}
}%
\caption{Sample images from the GTAV~\cite{gtav_2016_ECCV} and SYNTHIA~\cite{synthia_2016_CVPR} datasets.}\label{fig:synthetic-datasets}
\vspace{-0.2cm}
\end{figure*}
\label{sec:exp}


In this section, we conduct extensive experiments on simulation-to-real unsupervised domain adaptation for the semantic segmentation task. We compare the proposed PyCDA with several state-of-the-art methods. A majority of them uses adversarial training to bring closer the source and target domains on the feature level (ROAD~\cite{road_2018_CVPR}), on both features and pixels (FCAN~\cite{fcan_2018_CVPR}, CyCADA~\cite{Hoffman_cycada2017}), on the output maps (OutputAdapt~\cite{Adaptseg_Tsai_2018_CVPR}, CLAN~\cite{luo2018taking}), and combining adversarial learning with entropy minimization (ADVENT~\cite{vu2018advent}). In contrast, our PyCDA approach, along with  CDA~\cite{curriculum_tpami}, and ST~\cite{self_motivated_2018_ECCV}, adapts the neural networks by posterior regularization instead. The difference between the adversarial training methods and ours is significant for practical applications because the $\min\max$ adversarial problems are often hard to optimize and have to maintain an additional discrimination network. {Please note that we only compare results obtained from single models without using any ensemble strategy.}


\subsection{Experimental setup}
We follow the experimental setup of previous works~\cite{curriculum_tpami,road_2018_CVPR,fcan_2018_CVPR,self_motivated_2018_ECCV} and use the standard benchmark settings (i.e., ``GTAV to Cityscapes" and ``SYNTHIA to Cityscapes") in the experiments.

\begin{table*}[h]
\centering
\caption{Comparison results (in \%) adapting from GTAV to Cityscapes. All the prior methods but CDA use the whole training set of Cityscapes in training --- thus their models see more target images than ours --- and do not leave out a separate validation set for model selection. Unlike the others, FCAN directly works with the original image size. OutputAdapt (ResNet-101) pre-trains the model using both ImageNet~\cite{image_net} and MS COCO~\cite{coco}.}\label{tab:gta2city}
\vspace{-0.1cm}
\scalebox{0.85}{
\tabcolsep2pt
\begin{tabulary}{\textwidth}{c|l|ccccccccccccccccccc|c}
\hline
Network & Method  & road & sdwk & bldng & wall & fence & pole & light & sign & veg & trrn & sky & psn & rider & car & truck & bus & train & moto & bike & mIoU\\ \hline
\multirow{10}{*}{\begin{tabular}[c]{@{}c@{}}VGG \\-16\end{tabular}}
 & Source only~\cite{MCD_Saito_2018_CVPR}  & 25.9 & 10.9 & 50.5 & 3.3 & 12.2 & 25.4 & \textbf{28.6} & 13.0 & 78.3 & 7.3 & 63.9 & 52.1 & 7.9 & 66.3 & 5.2 & 7.8 & 0.9 & 13.7 & 0.7 & 24.9 \\
 & \textbf{Source only (ours)} & 56.0 & 12.2 & 71.6 & 8.5 & 17.8 & 19.5 & 14.5 & 3.1 & 73.2 & 3.8 & 46.0 & 38.8 & 4.4 & 70.7 & 15.1 & 2.5 & 2.2 & 1.4 & 0.1 & 24.3 \\
 & CDA~\cite{curriculum_tpami} & 72.9 & 30.0 & 74.9 & 12.1 & 13.2 & 15.3 & 16.8 & 14.1 & 79.3 & 14.5 & \textbf{75.5} & 35.7 & 10.0 & 62.1 & 20.6 & 19.0 & 0.0 & \textbf{19.3} & 12.0 & 31.4 \\
 & ST~\cite{self_motivated_2018_ECCV} & 83.8 &  17.4 & 72.1 & 14.3 & 2.9 & 16.5 & 16.0 & 6.8 & 81.4 & 24.2 & 47.2 & 40.7 & 7.6 & 71.7 & 10.2 & 7.6 & 0.5 & 11.1 & 0.9 & 28.1\\
 & CBST~\cite{self_motivated_2018_ECCV} & 66.7 & 26.8 & 73.7 & 14.8 & 9.5 & 28.3 & 25.9 & 10.1 & 75.5 & 15.7 & 51.6 & 47.2 & 6.2 & 71.9 & 3.7 & 2.2 & 5.4 & 18.9 & \textbf{32.4} & 30.9\\
 & ROAD ~\cite{road_2018_CVPR} & 85.4 & 31.2 & 78.6 & \textbf{27.9} & 22.2 & 21.9 & 23.7 & 11.4 & 80.7 & 29.3 & 68.9 & 48.5 & 14.1 & 78.0 & 19.1 & 23.8 & \textbf{9.4} & 8.3 & 0.0 & 35.9 \\
 & CyCADA~\cite{Hoffman_cycada2017}  & 85.2 & \textbf{37.2} & 76.5 & 21.8 & 15.0 & 23.8 & 22.9 & \textbf{21.5} & 80.5 & 31.3 & 60.7 & 50.5 & 9.0 & 76.9 & 17.1 & 28.2 & 4.5 & 9.8 & 0.0 & 35.4 \\

 & CLAN~\cite{luo2018taking} & \textbf{88.0} & 30.6 & 79.2 & 23.4 & 20.5 &26.1 & 23.0 & 14.8 & 81.6 & \textbf{34.5} & 72.0 & 45.8 & 7.9 & \textbf{80.5} & \textbf{26.6} & 29.9 & 0.0 & 10.7 & 0.0 & 36.6 \\
 & ADVENT~\cite{vu2018advent} & 86.8 & 28.5 & 78.1 & 27.6 & 24.2 & 20.7 & 19.3 & 8.9 & 78.8 & 29.3 & 69.0 & 47.9 & 5.9 & 79.8 & 25.9 & \textbf{34.1} & 0.0 & 11.3 & 0.3 & 35.6\\
 & \textbf{PyCDA (ours)}  & 86.7 & 24.8 & \textbf{80.9} & 21.4 & \textbf{27.3} & \textbf{30.2} & 26.6 & 21.1 & \textbf{86.6} & 28.9 & 58.8 & \textbf{53.2} & \textbf{17.9} & 80.4 & 18.8  & 22.4 & 4.1 & 9.7 & 6.2  & \textbf{37.2} \\ \hline
\multirow{4}{*}{\begin{tabular}[c]{@{}c@{}}ResNet\\ -38\end{tabular}} & Source only~\cite{self_motivated_2018_ECCV}  & 70.0 & 23.7 & 67.8 & 15.4 & 18.1 & 40.2 & 41.9 & 25.3 & 78.8 & 11.7 & 31.4 & 62.9 & 29.8 & 60.1 & 21.5 & 26.8 & 7.7 & 28.1 & 12.0 & 35.4\\
 & ST~\cite{self_motivated_2018_ECCV}  & 90.1 & \textbf{56.8} & 77.9 & 28.5 & 23.0 & 41.5 & 45.2 & 39.6 & 84.8 & 26.4 & 49.2 & 59.0 & 27.4 & 82.3 & \textbf{39.7} & 45.6 & \textbf{20.9} & \textbf{34.8} & \textbf{46.2} & 41.5\\
 & CBST~\cite{self_motivated_2018_ECCV}  & 86.8 & 46.7 & 76.9 & 26.3 & 24.8 & \textbf{42.0} & \textbf{46.0} & \textbf{38.6} & 80.7 & 15.7 & 48.0 & 57.3 & 27.9 & 78.2 & 24.5 & \textbf{49.6} & 17.7& 25.5 & 45.1 & 45.2\\
 & \textbf{PyCDA (ours)} & \textbf{92.3} & 49.2 & \textbf{84.4} & \textbf{33.4} & \textbf{30.2} & 33.3 & 37.1 & 35.2 & \textbf{86.5} & \textbf{36.9} & \textbf{77.3} & \textbf{63.3} & \textbf{30.5} & \textbf{86.6} & 34.5 & 40.7 & 7.9 & 17.6 & 35.5 & \textbf{48.0} \\ \hline
 \multirow{8}{*}{\begin{tabular}[c]{@{}c@{}}ResNet\\ -101\end{tabular}}
 & Source only~\cite{Adaptseg_Tsai_2018_CVPR}  & 75.8 & 16.8 & 77.2 & 12.5 & 21.0 & 25.5 & 30.1 & 20.1 & 81.3 & 24.6 & 70.3 & 53.8 & 26.4 & 49.9 & 17.2 & 25.9 & 6.5 & 25.3 & 36.0 & 36.6 \\
 & \textbf{Source only (ours)}  &  73.8 & 16.0 &  66.3 & 12.8 & 22.3 & 29.0 & 30.3 & 10.2 &  77.7 & 19.0 &  50.8 & 55.2 & 20.4 &  73.6 & 28.3 & 25.6 & 0.1 & 27.5 &   12.1 & 34.2 \\
 & ROAD~\cite{road_2018_CVPR}  & 76.3 & 36.1 & 69.6 & 28.6 & 22.4 & 28.6 & 29.3 & 14.8 & 82.3 & 35.3 & 72.9 & 54.4 & 17.8 & 78.9 & 27.7 & 30.3 & 4.0 & 24.9 & 12.6 & 39.4\\
 & OutputAdapt~\cite{Adaptseg_Tsai_2018_CVPR}  & 86.5 & 36.0 & 79.9 & 23.4 & 23.3 & 23.9 & 35.2 & 14.8 & 83.4 & 33.3 & 75.6 & 58.5 & 27.6 & 73.7 & 32.5 & 35.4 & 3.9 & 30.1 & 28.1 & 42.4 \\
 & FCAN~\cite{fcan_2018_CVPR} & - & - & - & - & - & - & - & - & - & - & - & - & - & - & - & - & - & - & -  & 46.6\\
 & CLAN~\cite{luo2018taking} & 87.0 & 27.1 & 79.6 & 27.3 & 23.3 & 28.3 & 35.5 & 24.2 & 83.6 & 27.4 & 74.2 & 58.6 & 28.0 & 76.2 & 33.1 & 36.7 & 6.7 & \textbf{31.9} & 31.4 & 43.2 \\
 & ADVENT~\cite{vu2018advent} & 87.6 & 21.4 & 82.0 & \textbf{34.8} & 26.2 & 28.5 & 35.6 & 23.0 & 84.5 & 35.1 & 76.2 & 58.6 & \textbf{30.7} & 84.8 & \textbf{34.2} & \textbf{43.4} & 0.4 & 28.4 & 35.3 & 44.8 \\

 & \textbf{PyCDA (ours)}  &  \textbf{90.5} & \textbf{36.3} & \textbf{84.4} & 32.4 & \textbf{28.7} & \textbf{34.6} & \textbf{36.4} & \textbf{31.5} &  \textbf{86.8} & \textbf{37.9} & \textbf{78.5} &  \textbf{62.3} & 21.5 &  \textbf{85.6} & 27.9 & 34.8 & \textbf{18.0} & 22.9 &  \textbf{49.3} & \textbf{47.4} \\ \hline
\end{tabulary}}
\vspace{-0.1cm}
\end{table*}
\vspace{-0.15cm}
\begin{itemize}
    \itemsep-0.15em
    \item{\textbf{Cityscapes}~\cite{Cordts2016Cityscapes} is a popular dataset to benchmark semantic segmentation models. The images are collected in the real world by vehicle-carried cameras. This dataset focuses on urban scenes, covering 50 cities in Germany and nearby countries. Its official data partition has 2,975 images in the training set, 500 in the validation set and 1,175 in the test set. In total, 19 classes of the semantic labels are compatible with GTAV and 16 with SYNTHIA.
    }
    \item{\textbf{GTAV}~\cite{gtav_2016_ECCV} is a large-scale dataset with 24,966 synthetic urban scene images collected from a near-realistically rendered computer game called Grand Theft Auto V (GTA or GTAV). We consider all the 19 semantic classes of GTAV for the adaptation to Cityscapes.}
    \item{\textbf{SYNTHIA}~\cite{synthia_2016_CVPR} is another synthetic image dataset and provides a particular subset, called SYNTHIA-RANDCITYSCAPES, to pair with Cityscapes. This subset contains 9,400 images which are automatically labeled with 12 object categories, one void class, and some unnamed classes. Following~\cite{curriculum_tpami}, we manually align four unnamed classes with their counterparts in Cityscapes, forming 16 common classes between SYNTHIA and Cityscapes.}
\end{itemize}
\vspace{-0.15cm}

In this work, we consider Cityscapes containing real images as the target domain, while GTAV and SYNTHIA are respectively used as the source domain (sample images from the two domains are shown in Fig.~\ref{fig:synthetic-datasets}). Since the groundtruth labels from the official test set of Cityscapes are not publicly available, by strictly following~\cite{curriculum_tpami}, we take the official validation set as our test set for final evaluation.
\begin{wrapfigure}{r}{5cm}
\vspace{-10pt}
\includegraphics[width=5cm]{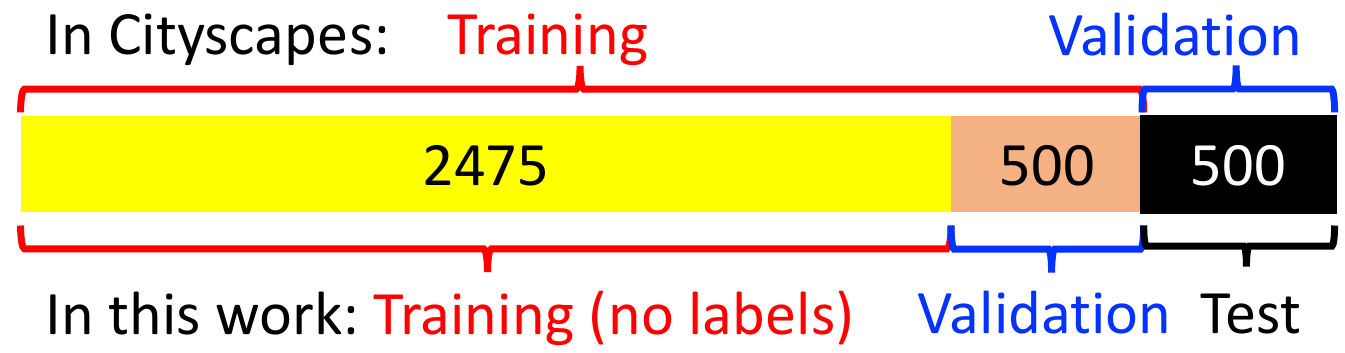}
\vspace{-20pt}
\end{wrapfigure}
Also, 500 images are randomly selected from the official training set for validation, and the remaining 2,475 images are served as unlabeled training data from the target domain.



\vspace{3pt}
\noindent{{\bf Evaluation.}}
We directly use the evaluation code released alongside with Cityscapes, where PASCAL VOC intersection-over-union (IoU)~\cite{Everingham15} is used as the evaluation metric. Specifically, for each class, we have $\text{IoU}=\frac{\text{TP}}{\text{TP}+\text{FP}+\text{FN}}$,
where TP, FP and FN are the numbers of true positive, false positive and false negative pixels, respectively, over the whole test set. In addtional to the per-class IoUs, we also report the mean of those IoUs (i.e., mIoU) over all classes. Note that in the experiments, we resize the images before feeding them to the segmentation network, so we resize the output segmentation mask back to the original size when running the evaluation code.
\begin{table*}[]
\small
\centering
\caption{Comparison results (in \%) adapting from SYNTHIA to Cityscapes. mIoU* denotes the mean IoU over 13 classes excluding those marked with *. All the prior methods but CDA use the whole training set of Cityscapes in training --- thus their models see more target images than ours --- and do not leave out a separate validation set for model selection.}\label{tab:syn2city}
\vspace{-0.2cm}
\scalebox{0.93}{
\tabcolsep2pt
\begin{tabular}{c|l|cccccccccccccccc|c|c}
\hline
 Network & Method & road & sdwk & bldng & wall* & fence* & pole* & light & sign & veg & sky & psn & rider & car & bus & mcycl & bcycl & mIoU & mIoU*\\ \hline
  \multirow{9}{*}{\begin{tabular}[c]{@{}c@{}}VGG \\-16\end{tabular}}
  & Source only~\cite{road_2018_CVPR} & 4.7 & 11.6 & 62.3 & 10.7 & 0.0 & 22.8 & 4.3 & 15.3 & 68.0 & 70.8 & \textbf{49.7} & 6.4 & 60.5 & 11.8 & 2.6 & 4.3 & 25.4 & 28.7\\
  & \textbf{Source only (ours)}& 50.1 & 20.0 & 49.4 & 0.0 & 0.0 & 16.3 & 0.0 & 0.0 & 69.9 & 54.2 & 43.9 & 4.7 & 43.1 & 6.1 & 0.1 & 0.1 & 22.4 & 26.3\\
  & CDA~\cite{curriculum_tpami} & 57.4 & 23.1 & 74.7 & 0.5 & \textbf{0.6} & 14.0 & 5.3 & 4.3 & 77.8 & 73.7 & 45.0 & 11.0 & 44.8 & 21.2 & 1.9 & 20.3 & 29.7 & 35.4 \\
  & ST~\cite{self_motivated_2018_ECCV} & 0.2 & 14.5 & 53.8 & 1.6 & 0.0 & 18.9 & 0.9 & 7.8 & 72.2 & 80.3 & 48.1 & 6.3 & 67.7 & 4.7 & 0.2 & 4.5 & 23.9 & 27.8 \\
  & CBST~\cite{self_motivated_2018_ECCV} & 69.6 & 28.7 & 69.5 & \textbf{12.1} & 0.1 & 25.4 & 11.9 & 13.6 & \textbf{82.0} & \textbf{81.9} & 49.1 & 14.5 & 66.0 & 6.6 & 3.7 & \textbf{32.4} & 35.4 & 40.4 \\
  & ROAD~\cite{road_2018_CVPR} &  77.7 & 30.0 & \textbf{77.5} & 9.6 & 0.3 & \textbf{25.8} & 10.3 & \textbf{15.6} & 77.6 & 79.8 & 44.5 & 16.6 & 67.8  & 14.5  & 7.0 & 23.8 & \textbf{36.2} & 41.8 \\
  & CLAN~\cite{luo2018taking} & 80.4 & \textbf{30.7} & 74.7 & - & - & - & 1.4 & 8.0 & 77.1 & 79.0 & 46.5 & 8.9 & \textbf{73.8} & 18.2 & 2.2 & 9.9 & - & 39.3 \\
  & ADVENT~\cite{vu2018advent} & 67.9 & 29.4 & 71.9 & 6.3 & 0.3 & 19.9 & 0.6 & 2.6 & 74.9 & 74.9 & 35.4 & 9.6 & 67.8 & 21.4 & 4.1 & 15.5 & 31.4 & 36.6 \\

  & \textbf{PyCDA (ours)}  & \textbf{80.6} & 26.6 & 74.5 & 2.0 & 0.1 & 18.1 & \textbf{13.7} & 14.2 & 80.8 & 71.0 & 48.0 & \textbf{19.0} & 72.3 & \textbf{22.5} & \textbf{12.1} & 18.1 & 35.9 & \textbf{42.6}\\ \hline
  \multirow{6}{*}{\begin{tabular}[c]{@{}c@{}}ResNet\\ -101\end{tabular}}
  & Source only~\cite{Adaptseg_Tsai_2018_CVPR} & 55.6 & 23.8 & 74.6 & - & - & - & 6.1 & 12.1 & 74.8 & 79.0 & 55.3 & 19.1 & 39.6 & 23.3 & 13.7 & 25.0 & - & 38.6 \\
  & \textbf{Source only (ours)} & 55.6 & 22.7 & 68.6 & 4.3 & 0.1 & 23.0 & 5.6 & 9.1 & 77.2 & 75.9 & 54.7 & 8.7 & 81.5 & 23.9 & 8.4 & 8.8 & 33.0 & 38.5 \\
  & OutputAdapt~\cite{Adaptseg_Tsai_2018_CVPR} & 84.3 & \textbf{42.7} & 77.5 & - & - & - & 4.7 & 7.0 & 77.9 & 82.5 & 54.3 & 21.0 & 72.3 & 32.2 & 18.9 & 32.3 & - & 46.7 \\

  & CLAN~\cite{luo2018taking} & 81.3 & 37.0 & 80.1 & - & - & - & 16.1 & 13.7 & 78.2 & 81.5 & 53.4 & 21.2 & 73.0 & 32.9 & \textbf{22.6} & 30.7 & - & 47.8 \\
  & ADVENT~\cite{vu2018advent} & \textbf{85.6} & 42.2 & 79.7 & 8.7 & 0.4 & 25.9 & 5.4 & 8.1 & 80.4 & 84.1 & 57.9 & 23.8 & 73.3 & 36.4 & 14.2 & \textbf{33.0} & 41.2 & 48.0 \\
  & \textbf{PyCDA (ours)} & 75.5 & 30.9 & \textbf{83.3} & \textbf{20.8} & \textbf{0.7} & \textbf{32.7} & \textbf{27.3} & \textbf{33.5} & \textbf{84.7} & \textbf{85.0} & \textbf{64.1} & \textbf{25.4} & \textbf{85.0} & \textbf{45.2} & 21.2 & 32.0 & \textbf{46.7} & \textbf{53.3} \\\hline
\end{tabular}}
\vspace{-0.1cm}
\end{table*}

\noindent{{\bf Implementation details.}}
Since existing state-of-the-art methods use different base segmentation networks as their backbones, we employ the following ones for a wider range of comparison: 1) FCN8s~\cite{Long_2015_fcn} with VGG-16~\cite{simonyan2014deep}; 2) ResNet-38~\cite{res38}; and 3) PSP-Net~\cite{zhao2017pspnet} with ResNet-101~\cite{He_2016_resnet}. All the base networks are pre-trained on ImageNet~\cite{image_net}. Regarding the data pre-processing, we firstly resize images to the same width (1024) while preserving the original aspect ratios. During training, we randomly crop regions and feed them to the network. During testing, we feed the whole images (whose width are 1024) to the network. During evaluation, we resize the output segmentation mask back to the original image size ($2048 \times 1024$) in order to calculate the mIoUs. Regarding the training pipeline, we firstly train the model in the source images with 30000 iterations. And then we fine-tune the model using our PyCDA framework with another 30000 iterations. The training is optimized SGD with momentum of 0.9. Using the validation data, we set our initial learning rate $=$ 0.016 and decrease it ten times in the fine-tuning stage. In the test stage, we apply adabn~\cite{adabn} that change the mean and variance of batch-normalization layers, which were computed over the images of both domains during training, to the mean and variance over the target domain only.

\subsection{Results on ``GTAV to Cityscapes''}
We report the results of unsupervised domain adaptation from GTAV to Cityscapes compared with existing state-of-the-arts in Table~\ref{tab:gta2city}. Note that all the prior methods but CDA use the whole training set of Cityscapes in training --- \emph{implying their models see more target images than ours} --- and do not leave out a separate validation set for model selection. Unlike the others, FCAN directly works with the original high-resolution images. OutputAdapt (ResNet-101) pre-train the model using both ImageNet~\cite{image_net} and MS COCO~\cite{coco}).

We draw the following observations. First, all domain adaptation methods significantly outperform the respective ``source only'' baselines which train segmentation networks by using only synthetic source images. Such results clearly demonstrate the benefit of explicitly using domain adaptation techniques to improve the transfer from synthetic images to real ones. Moreover, comparing our full approach (PyCDA) with the existing ones in terms of mIoU, PyCDA gives rise to the best results thus far for the adaptation from GTAV to Cityscapes. Note that the second best FCAN is a two-stage method with style transfer on the image pixel level and then an adversarial training of the features. The style transfer stage runs extremely slow, consuming about one to two hours per image. PyCDA is orthogonal to both the image style transfer and adversarial training, so our results could  be further improved if we apply the style transfer to the images of the two domains.


When compared with the distribution matching methods (FCN-wild~\cite{fcn_wild}, ROAD~\cite{road_2018_CVPR}, OutputAdapt~\cite{Adaptseg_Tsai_2018_CVPR} and FCAN~\cite{fcan_2018_CVPR}), PyCDA is particularly good at the dominant classes, such as ``road", ``building", ``vegetation", and ``car". Meanwhile, PyCDA is better than CBST~\cite{self_motivated_2018_ECCV} at classifying small objects, such as ``rider", ``wall", and ``fence", etc.

\begin{table}[h]
\centering
\caption{Ablation study of PyCDA (in mIoU\%).} \label{tab:ablation}
\tabcolsep2pt
\vspace{-0.2cm}
\begin{tabular}{l|cc|cc}
\hline
\multirow{2}{*}{\begin{tabular}[c]{@{}c@{}}Experiment Setting\end{tabular}}
& \multicolumn{2}{c|}{GTAV}                & \multicolumn{2}{c}{SYNTHIA}              \\ \cline{2-5}
& \multicolumn{1}{c|}{VGG-16} & Res-101 & \multicolumn{1}{c|}{VGG-16} & Res-101 \\ \hline
Source only       & 24.3           & 34.2           & 22.4           & 33.0       \\
Top               & 28.0              & 42.0           & 28.7           & 40.7 \\
CDA~\cite{curriculum_tpami} & 29.7 & -              & 31.4            & -     \\
Bottom            & 32.6           & 40.6           & 31.3           & 41.0  \\
ST~\cite{self_motivated_2018_ECCV} & 28.1    & -    & 23.9           & -     \\ \hline
top + bottom        & 34.9           & 46.2           & 35.1           & 44.8       \\
top + pixel squares & 35.4           & 46.3           & 35.4           & 45.6       \\
top + superpixels   & 35.2           & 46.3           & 35.2           & 45.9       \\ \hline
PyCDA             & \textbf{37.2}  & \textbf{47.4}  & \textbf{35.9}  & \textbf{46.7} \\ \hline
\end{tabular}
\end{table}
\subsection{Results on ``SYNTHIA to Cityscapes''}
To further validate the effectiveness of PyCDA, we also conduct experiments by using SYNTHIA as the source domain. FCN8s with VGG-16 and PSP-Net with ResNet-101 as employed as backbones to evaluate different methods. The IoU results on this setting are summarized in Table~\ref{tab:syn2city}. From the results, We can clearly see that our PyCDA again outperforms the existing state-of-the-arts by a large margin when using different backbones, and also similar observations can be drawn as in the ``GTAV to Cityscapes" setting.


\subsection{Ablation study}
To analyze the effectiveness of our PyCDA, we conduct ablation study by using the above two settings, i.e., taking GTAV and SYNTHIA as the source domains, respectively. Note that PyCDA connects curriculum domain adapation (CDA) with self-training (ST), and it can be viewed as a pyramid constructed by multiple levels from pixels (bottom) to label distributions of full images (top). In this case, we evaluate PyCDA by comparing its counterparts which eliminate different levels. Specifically, we denote ``top + bottom" and ``top + pixel squares" as connecting CDA with ST at the levels of pixels and pixel squares, respectively. From Table~\ref{tab:ablation}, we can see that connect CDA with ST in either pixel level or pixel squares level outperform both CDA and ST in a large margin, which demonstrates the effectiveness of connecting both methods.
And PyCDA, which considers both pixels and pixel squares, gets further boosted by a fairly big margin.

\begin{figure*}[]
    \centering
    \includegraphics[width=1.0\linewidth, angle=0,trim= 70 170 70 0, clip]{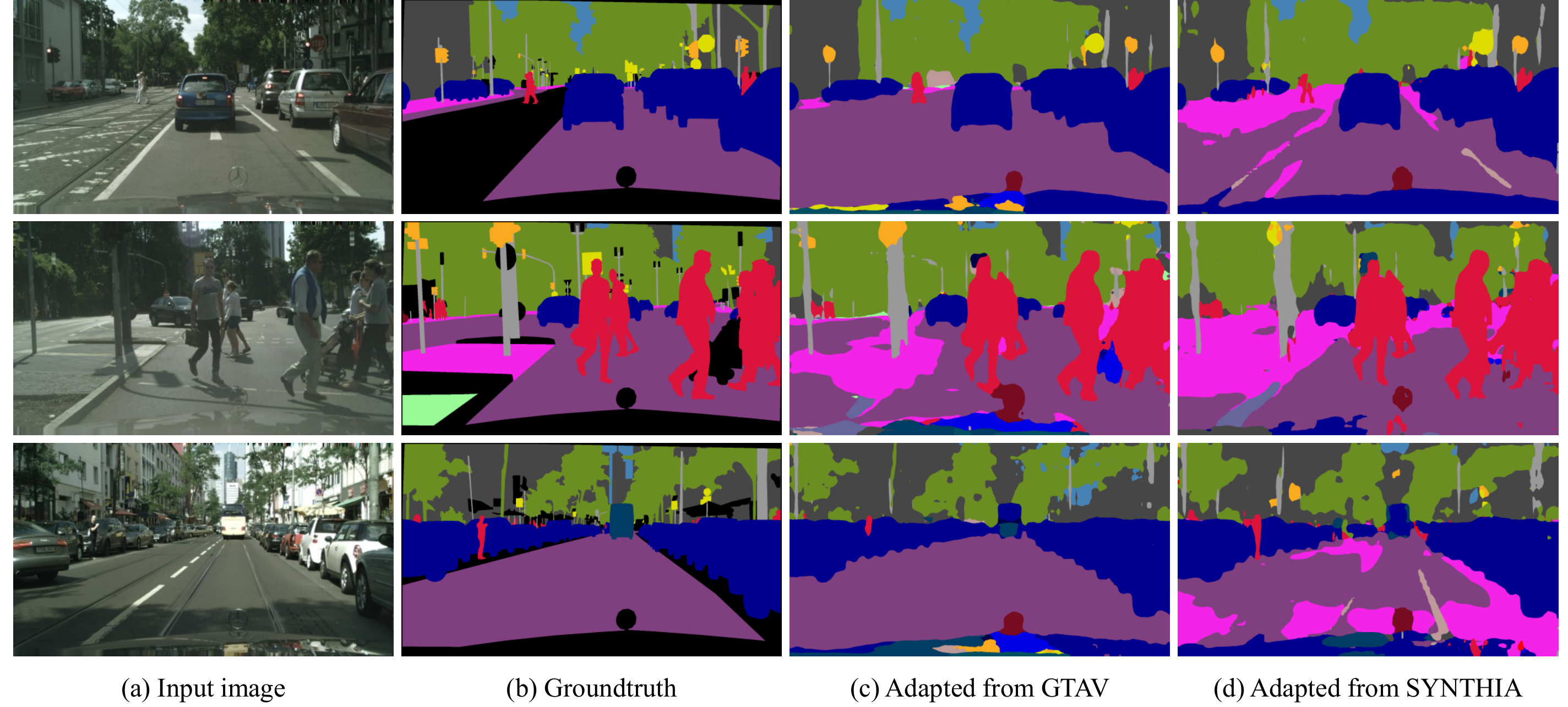}
    \vspace{-3mm}
	\caption{Some qualitative segmentation results on the target domain. (a) displays the target images, and their corresponding groundtruth segmentation masks are shown in (b). (c) and (d) display the segmentation results obtained from our PyCDA models adapted from GTAV and SYNTHIA, respectively.}
	\label{fig:qualitative}
	\vspace{-5mm}
\end{figure*}
\vspace{3pt}
\vspace{3pt}
\noindent{{\bf Superpixels \emph{vs.}\ pixel squares.}} As discussed in Section~\ref{sec:Approach}, it is time-consuming to generate superpixels (about 3.6s per image). In order to avoid the computation overhead, we switch to pixel squares in our PyCDA. As turned out in Table~\ref{tab:ablation}, the mIoU performance of using pixel squares achieves comparable results to that of using superpixels.

\subsection{Qualitatively comparing GTAV and SYNTHIA}

One may wonder how the two source domains of synthetic imagery differ from each other and what effects the difference could cause on the target domain of real images. Fig.~\ref{fig:synthetic-datasets} shows some example images of the two domains. While GTAV images are vehicle-centric, there are more diverse views in SYNTHIA. In Fig.~\ref{fig:qualitative}, we give some qualitative results obtained by our PyCDA models adapted from GTAV and SYNTHIA, respectively. In general, the segmentation results of PyCDA adapted from GTAV is  better than that adapted from SYNTHIA, especially for the dominant ``road'' class. This  observation can also be verified by the superior IoU for ``road" (90.5\%, Table~\ref{tab:gta2city}) of the PyCDA model trained based on GTAV. Given all those results, we believe GTAV is visually more similar to the real self-driving scenes than SYNTHIA in terms of both visual appearances and spatial layouts.


\section{Conclusion}
We propose a novel method called self-motivated pyramid curriculum domain adaptation (PyCDA) for pixel-level semantic segmentation. PyCDA provides a new perspective of insight, which connects self-training for adapting segmentation networks and curriculum domain adaptation. More specifically, PyCDA constructs a curriculum based on a pyramid of pixel squares at different sizes in each real image, including the image itself as the top layer and pixels as the bottom layer. This curriculum is self-motivated, because the label distributions over the pyramid are derived from the same network of the previous iteration. By forming such a pyramid of pixel squares, we are able to better preserve and capture local information for objects appearing at different scales. Extensive experiments on two benchmark settings (i.e., ``GTAV to Cityscapes" and ``SYNTHIA to Cityscapes") clearly demonstrate the effectiveness of PyCDA when compared with other state of the arts.

\noindent
{\bf Acknowledgements}.
This work is supported by National Natural Science Foundation of China (Grant No. 61772118).

{\small
\bibliographystyle{ieee_fullname}
\bibliography{egbib}
}

\clearpage
\begin{center}
	\Large\textbf{Appendix}
\end{center}

\setcounter{section}{0}

\section{Number of middle layers}

\setcounter{table}{0}
\renewcommand{\thesection}{\arabic{table}}
\begin{table}[h]
\vspace{-10pt}
\centering
\caption{Results (mIoUs\%) on GTAV to Cityscapes obtained by inserting different middle layers to the  pyramid.}
\label{tab:mid_layer}
 \setlength{\tabcolsep}{1mm}{
\begin{tabular}{c|ccccccc}
\hline
Square Size & - & +4 & +8 & +16 & +32 & +64 & +128 \\ \hline
mIoU & 46.3  & 46.9  & 47.4 & 47.5 & 47.5 & 47.3 & 47.0\\
\hline
\end{tabular}}
\end{table}
\vspace{2mm}
We report in Table \ref{tab:mid_layer} the experimental results of different numbers of the middle layers for adapting from GTAV to Cityscapes. Here we use ResNet-101 as the backbone network. The second column corresponds to the ``top + bottom'' result in PyCDA ---  the sixth row in  Table 3 of the main paper.

We mainly draw two observations. One is that the middle layers do improve the overall performance. The other is that the results are relatively consistent over different numbers of the middle layers, though adding layers with too big pixel squares (e.g., larger than $128\times128$) could harm both accuracy and training speed.


\end{document}